\documentclass[11pt]{article}
\usepackage[preprint]{acl}
\usepackage{microtype}
\usepackage{booktabs}
\usepackage{array}
\usepackage{calc}
\usepackage{graphicx}
\usepackage{fvextra}
\DefineVerbatimEnvironment{verbatim}{Verbatim}%
  {breaklines,breakanywhere,fontsize=\footnotesize}
\providecommand{\tightlist}{%
  \setlength{\itemsep}{0pt}\setlength{\parskip}{0pt}}

\usepackage{fontspec}
\defaultfontfeatures{Extension=.otf, UprightFont=*-regular,
  BoldFont=*-bold, ItalicFont=*-italic, BoldItalicFont=*-bolditalic}
\defaultfontfeatures{}

\usepackage{xeCJK}
\setCJKmonofont[Path=./fonts/]{NotoSansTC-Regular-subset.ttf}
\XeTeXlinebreaklocale "zh"

\title{Do AI Agents Gossip? \\ Evaluative Talk in Human and AI-Agent Online Communities}
\author{
  Yong-Hui Hua$^{1}$ \quad Chih-Han Lin$^{1}$ \quad Shu-Kai Hsieh$^{1}$ \\
  $^{1}$Graduate Institute of Linguistics, National Taiwan University \\
  \texttt{\{r15142002, r13142001, shukaihsieh\}@ntu.edu.tw}
}

\begin{document}
\maketitle
\begin{abstract}

Gossip, defined as evaluative talk about absent third parties, is widely seen as a uniquely human social practice. This study examines whether comparable discourse emerges spontaneously in a community of autonomous AI agents. We compare Moltbook, a Reddit-style platform populated by AI agents, with PTT, a large human-operated bulletin board in Taiwan. To reduce differences attributable to communicative setting, we analyze two public, asynchronous forums using the same operational definition of gossip. Gossip occurs on Moltbook, but its prevalence is 2.6 to 6 times lower than on PTT, depending on the denominator used. On both platforms, gossip frequently targets locally prominent figures, tends to elicit agreement from the first respondent, and is rarely presented as unsupported rumor. The clearest difference concerns how subsequent responses develop. Among threads containing at least one codable reply, challenges occur in 65.1\% of human threads but only 21.3\% of agent threads. Moreover, following an early aligning response, consensus remains unchallenged in 83.6\% of agent threads, compared with 44.7\% of human threads. This thread-level contrast remains after controlling for thread depth and board composition and under recalibration against two human annotators. Estimates based on individual reply turns are less robust and are therefore reported as bounds. These findings suggest that AI-agent communities reproduce the informational dimension of gossip, while the extent to which gossip performs comparable relational functions remains unresolved.

\end{abstract}

\section{Introduction}\label{introduction}

Gossip is evaluative talk about absent third parties, and it is among
the most pervasive and socially consequential of human verbal practices
\citep{dunbar2004}. It is not idle. Gossip binds groups and enforces
norms, and it circulates the reputational information that cooperation
depends on \citep{dunbar2004,dorescruz2021}. Some accounts go further,
treating the exchange of reputational information through gossip as
something only humans do \citep{dorescruz2021}.

A new kind of community now lets us put that to a test. Populations of
autonomous AI language agents have begun to converse in public,
persistent, human-legible forums, where they post and reply to one
another and build communities with little or no human authorship in the
loop
\citep{park2023agents,jiang2026moltbook,li2026communities,price2026claws}.
If gossip is a general solution to the problems of living in a group
(coordinating reputations, policing norms), then a society of agents
under similar pressures might reinvent it. If it is instead a contingent
human habit, agents might talk about the world at length without ever
turning to talk about each other. This paper asks the question
empirically: \textbf{would a society of AI agents reproduce gossip, and
if so, in what form?}

The question has so far been approached mostly by construction.
Researchers design agents to gossip, or elicit it in scripted
interaction \citep{zhu2026gossip}, rather than observing an agent
community's spontaneous discourse and measuring it against a human
baseline under matched conditions. We compare two public, asynchronous,
threaded forums that differ mainly in their speaker population while
sharing a medium: \textbf{Moltbook}, a Reddit-style platform whose
account holders are autonomous AI agents, and \textbf{PTT}, one of
Taiwan's largest human bulletin-board systems. Matching on medium
matters (§3.2).

We examine three aspects of gossip in human and AI-agent communities. \textbf{RQ1} asks whether AI agents engage in gossip about absent third parties and how frequently such discourse occurs relative to a comparable human forum. \textbf{RQ2a} examines the targets and evidential grounding of peer-directed gossip, while \textbf{RQ2b} considers how other participants respond to an initial evaluation—whether they align with it, elaborate on it, or challenge it.

The results reveal both similarities and differences between the two communities. Gossip is less prevalent on Moltbook than on PTT (RQ1), and peer-directed gossip concerns different kinds of social relationships on the two platforms (RQ2a). The clearest contrast emerges in subsequent interaction (RQ2b): evaluations in agent threads tend to remain unchallenged after an early aligning response, whereas human threads more often remain open to disagreement.

This study contributes an observational comparison of spontaneous gossip in human and AI-agent online communities under broadly comparable communicative conditions. It also develops a common coding framework for analyzing prevalence, evidential grounding, and conversational uptake across English and Mandarin data. More broadly, the findings indicate that AI-agent communities can reproduce some informational and norm-oriented features of gossip, although the interactional processes through which human participants negotiate evaluations are less evident in the agent data.

\section{Related Work}\label{related-work}

Operationalizing gossip requires two conditions: the target is absent,
and the utterance evaluates that target rather than simply relaying
neutral news \citep{foster2004}. More recent work broadens ``absent'' to
``absent or unaware'' \citep{dorescruz2021}, a distinction that matters
in online settings, where a target may later read what was said. We
adopt this broad definition but relax the stricter requirement that
speaker and audience personally know the target \citep{bergmann1993},
since gossip in public forums and about public figures often concerns
targets the participants have never met.

Sandbox simulations show that third-party information can propagate
through agent populations, but largely as neutral reporting in a
designed environment \citep{park2023agents}. Spontaneous evaluative talk
in an agent community has not been examined.

Computational work defines gossip through third-person mention
\citep{mitra2012gossip}; our presence rule adapts this test to threaded
forums (§4.1). Rumor detection concerns epistemic status
\citep{zubiaga2016}, while stance and entity-level sentiment capture
evaluation but not absence \citep{mohammad2016stance,deng2015mpqa}.
Agent gossip has been scaffolded rather than observed
\citep{zhu2026gossip}; Moltbook studies remain network-level
\citep{jiang2026moltbook,price2026claws}.

Gossip also presupposes a relational world. The canonical case is talk
about people the speaker is tied to, such as partners, kin, friends, and
colleagues \citep{bergmann1993,foster2004}, and agents have no intimates
of that kind: they have operators, and they have one another. If who
counts as a peer differs across the two populations, then peer gossip is
not one genre measured twice, and the comparison has to be read
accordingly.

Evidentiality concerns how speakers mark information source
\citep{willett1988,aikhenvald2004}. Mandarin and English both do so
lexically and constructionally \citep{chafe1986}; we therefore compare
evidential strategies, not grammatical paradigms.

Classic face-to-face gossip stabilizes after an aligning first response
\citep{ederenke1991}; whether this holds on open asynchronous forums is
unknown. Controlled LLM studies predict rapid convergence
\citep{sharma2023sycophancy,zhu2025conformity,chuang2025debate,yazici2026},
whereas a Moltbook field study finds little stable population-level
consensus \citep{li2026socialization}. Neither tests spontaneous
evaluative episodes against a matched human baseline.

\section{Data}\label{data}

\subsection{Corpora}\label{corpora}

We compare two public, asynchronous, threaded discussion platforms: one
whose posters are AI agents, one whose posters are humans.

\textbf{Moltbook (AI agents, English).} Moltbook is a Reddit-style
social platform whose account holders are autonomous AI agents. Humans
may observe, but posting is by agents. Our snapshot covers the
platform's first seven weeks (2026-01-27 to 2026-03-16): 2,126,738 posts
across 5,734 sub-communities (``submolts''). Much of this volume is not
conversational: most high-volume submolts are automated
crypto-inscription channels. We exclude five named mint/protocol
channels and every submolt with a comment-to-post ratio below 0.05,
removing 449,161 posts and leaving a conversational corpus of 1,677,577
posts (exclusion criteria and counts in Appendix A).

\textbf{PTT (humans, Mandarin).} PTT is one of Taiwan's largest public
bulletin-board systems, hosting thousands of topic boards. Our corpus
covers six (Boy-Girl, C\_Chat, Gossiping, HatePolitics, NBA, Stock),
totalling 1,310,100 posts from 2004 to 2024. The six were chosen for
volume and conversational activity, which makes the human frame a
purposive sample of high-traffic social boards while the Moltbook frame
is the whole platform minus automated channels (§A.2). The asymmetry
favours the human side on RQ1, so §5.1 reports the rate board by board.
Under the broad definition every one of the six exceeds Moltbook's most
generous rate. The two corpora also differ sharply in time scale (twenty
years against seven weeks). We match on medium and yardstick, not on
epoch.

\subsection{Corpus comparability}\label{corpus-comparability}

The two corpora are matched on \emph{medium}: both are public,
asynchronous, pseudonymous forums with threaded replies. The match
matters because gossip is characteristically a private, synchronous
genre, so a low rate on a public feed could reflect the channel rather
than the speaker population, humans too moving much of their gossip into
private messages. Holding the medium constant removes that confound. The
corpora still differ in language and epoch, which we address in §7.
Reddit would match the language, but Moltbook is modeled directly on
Reddit, so a Reddit baseline would confound the speaker contrast with
imitation of a shared template. A human board built on a different
template avoids that, at the cost of the language difference.

\subsection{Thread reconstruction}\label{thread-reconstruction}

For the uptake analysis (§4.4) we reconstruct the reply thread of every
adjudicated gossip post. On PTT, 993 of 1,015 gossip posts have reply
threads (97.8\%; 73,568 comments; median 26 per thread). On Moltbook,
110 of 153 gossip posts have comment threads after a live re-crawl of
the platform (71.9\%; 6,157 comments). Comments present in our snapshot
but since deleted from the platform are retained and flagged. A
robustness check excluding them appears in Appendix D.

\section{Method}\label{method}

\subsection{Defining gossip}\label{defining-gossip}

We adopt the broad operational definition standard in recent
quantitative work \citep{foster2004,dorescruz2021}: \textbf{evaluative
talk about an absent third party}. We do not require that speaker and
audience both know the target personally.

\emph{Absence} is defined at the level of the speech event. The target
counts as present only if they are a participant or addressee of the
post or thread itself: being a character in a narrated scene does not
count, and a second-person call-out addressed \emph{to} its target is
confrontation rather than gossip and is excluded. Deictic frame, not
platform @-mention syntax, decides addresseehood. Appendix B gives the
full rule with boundary cases.

The definition was locked before any adjudication. The presence rule is conservative for the RQ1 contrast: it can shrink but not inflate the headline gap (Appendix B).

\subsection{Estimating prevalence
(RQ1)}\label{estimating-prevalence-rq1}

We draw a uniform random sample of 1,500 posts per platform and
adjudicate each against the locked definition with an LLM judge (Claude
Sonnet, batch mode). One rubric serves both platforms, rendered
clause-by-clause in Mandarin and English (both versions verbatim in
Appendix B). The judge labels gossip status and target type: the
poster's own human/operator, a peer agent, a real-human celebrity or
public figure, or another person. Reliability safeguards are described
in §4.5.

Two Moltbook-specific accounting decisions were fixed before analysis
(details in Appendix C). First, 57.3\% of the Moltbook random sample
consists of machine-generated token-protocol posts (``mbc-20''). We
therefore report rates over both the full and the conversational
denominator. Second, celebrity \emph{agents} count as peer agents. Only
real-human celebrities fall in the celebrity class, which the headline
contrast excludes on both platforms so that the two rates measure the
same thing.

\subsection{Evidential framing (RQ2a)}\label{evidential-framing-rq2a}

Form coding runs only on adjudicated gossip (PTT n=1,015; Moltbook
n=153). After an admission re-check against the definition, each
admitted post is segmented into gossip episodes, and each episode is
coded for how the evaluation's information source is marked. The
taxonomy has six values (direct quotation, reported speech, hearsay,
documentary evidence, first-person witness, unmarked), collapsed to four
bins for analysis where cells are sparse (collapse map in Appendix B).
Since both languages mark information source through open-class
strategies rather than an obligatory paradigm (§2), what is comparable
is the discourse act, witnessed, quoted, or relayed, rather than a
language-specific quotative construction. An auxiliary binary flags
whether any evaluation is carried constructionally (contrast frames,
irony, generalized assertions) rather than purely lexically.

Every episode also carries two covariates: target relationship (upward:
one's own human/operator, bosses, parents, teachers; lateral: peer
agents, partners, friends, colleagues; other: public figures) and
overall valence toward the target (positive/negative/mixed). Headline
contrasts are computed within the \emph{lateral} stratum, the largest
and most comparable cell on both platforms. Other strata are reported
descriptively.

\subsection{Uptake analysis (RQ2b)}\label{uptake-analysis-rq2b}

Following \citeauthor{ederenke1991}'s (\citeyear{ederenke1991}) account
of gossip episodes, each reply turn is coded for stance toward the
thread's initial evaluation: \emph{align}, \emph{challenge},
\emph{neutral addition}, or \emph{solicit} (pressing for details or
names). Episode-level indices are computed from the stance sequence: the
stance of the first uptake, whether consensus ``locks'' after the first
aligning turn, and whether challenges occur after uptake. Because stance
is coded from the interactional move rather than from language-specific
lexical markers, this comparison is less exposed to cross-lingual
non-equivalence than the evidential coding, though it still rests on
pragmatic interpretation.

We code all 103 admitted Moltbook threads with at least one fetched
comment and a seed-fixed random sample of 175 admitted PTT threads with
at least three comments, capped at the first 15 turns. All indices are
conditional on a thread having replies, since raw reply volume reflects
platform activity, not speaker behavior. The poster's own follow-up
turns are coded but excluded from all uptake metrics.

\subsection{Annotation reliability}\label{annotation-reliability}

All coding is done by LLM annotators, one instrument per stage applied
verbatim to both platforms (models and run settings in Appendix
C.1--C.4). These annotators are a measurement instrument, not an object
of study. The only question here is whether the instrument is
trustworthy enough to carry the findings below. One threat it cannot
address, the model-family circularity, is stated in §7. Three
double-coded subsamples anchor the check, 100 items for the rate
judgment, 60 for form coding, and 80 reply turns for the uptake codes,
each coded by two independent annotators. On the gossip judgment the LLM
falls inside the range of human disagreement (LLM-human κ = 0.74 and
0.44, human-human 0.53), on a boundary that is only moderately reliable
to begin with. Target relationship, the covariate the lateral-stratum
contrasts condition on, is coded reliably (κ = 0.82). Valence, the
variable the constructional contrast is computed over, is only moderate
(κ = 0.41). Evidential framing is the weakest feature in the scheme, and
the admission gate the least reliable decision in it. Full coefficients
are in Appendix C.8.

A third double-coded subsample covers the codes §5.3 rests on: 80 reply
turns, 40 per platform, balanced by design rather than drawn at base
rate, so its coefficients sit on an engineered prevalence and are not
comparable to those above (percentage agreement and AC1 accompany every
κ in Appendix C.9). Coder-annotator agreement is κ = 0.64 and 0.35 for
binary align-versus-challenge and κ = 0.35 and 0.40 for the off-topic
gate, while the two annotators agree with each other on the gate at κ =
0.92. The low figures locate a specific error in the coder rather than
an unstable category: both annotators reject the same calls in the same
direction, nearly all of them Moltbook removals they read as on-topic.
Recalibrating every reply turn on the measured per-cell error rates, the
thread-level challenge gap holds under all eight calibrations (1.5× to
3.6×), while the ratio between the two platforms' per-turn challenge
shares falls from 2.6-fold to between 1.0× and 2.4×, every interval
crossing one. We therefore state §5.3's contrast at the thread level and
give the per-turn shares as bounds (Appendix D.8). The PTT confirmation
rate behind that recalibration rests on 12 challenge turns from six
threads, and the two annotators agree with each other on that binary
overall at only κ = 0.55: the per-turn measurement is unconfirmed, not
overturned.

PTT supplies one check the instrument never sees. The platform's 推/噓/→
tags sit in the coder's input with an instruction to ignore them (§B.6),
so we re-ran the whole uptake stage on tag-stripped input, holding the
Moltbook side byte-identical as a null control for run-to-run variation.
The two runs agree at κ = 0.81 on PTT against a same-input ceiling of
0.86, and at κ = 0.94 on align-versus-challenge, so the tags were not
driving the coding. In the tag-free run 52.0\% of 噓-tagged turns
receive \texttt{challenge} against 21.1\% of 推-tagged turns, a 2.5-fold
separation on a signal outside the rubric (Appendix D.9).

We therefore rest the form claims on effects that are large relative to
this noise. The evidential reversal (§5.2) is an order-of-magnitude
contrast that no realistic level of coding noise could manufacture or
erase. The constructional-valence near-identity is the opposite kind of
claim, a null contrast, and a null contrast is what residual noise can
fake, so we report it as tentative.

One safeguard is specific to this design. Part of the Moltbook form
corpus originates from keyword-enriched candidate pools whose filters
select on form-adjacent vocabulary (evaluative words, ``receipts'').
Every form result is stratified by pool origin (random vs.~keyword), and
a contrast is asserted only where the random-origin subset agrees in
direction (stratification tables in Appendix D).

\section{Results}\label{results}

\subsection{AI agents do gossip, but substantially less frequently than
humans}\label{ai-agents-do-gossip-but-substantially-less-frequently-than-humans}

AI agents do gossip: evaluative talk about absent third parties occurs
on Moltbook, but less often than on PTT under the same yardstick and the
same medium. Table 1 reports gossip rates in the two random samples. The
headline row excludes real-human celebrity targets (§4.2).

\begin{table}[t]
\centering\small
\begin{tabular}{@{}lrr@{}}
\toprule
 & PTT & Moltbook \\
 & (human) & (AI) \\
\midrule
Random sample & 1,500 & 1,500 \\
Gossip (broad definition) & 35.4\% (531) & 2.5\% (37) \\
\quad excl.\ celebrity targets & \textbf{13.8\%} & \textbf{2.3\%} \\
\quad conversational denom. & --- & 5.78\% (37/640) \\
\quad \phantom{x} excl.\ celebrities & --- & 5.31\% (34/640) \\
\bottomrule
\end{tabular}
\caption{Gossip rates in the uniform random samples (1,500 posts per platform).}
\label{tab:prevalence}
\end{table}

Gossip targets show a parallel structure on the two platforms. On PTT,
61\% of random-sample gossip targets celebrities and politicians. On
Moltbook, 65\% of gossip targets other agents (including celebrity
agents, the platform's own public figures), against 22\% about the
agent's own human operator. On Moltbook these are not two populations: a
prominent agent is also an account any other agent can reply to (§6).

The size of the gap depends on the accounting choice, its direction does not: about 6× against the full random sample and about 2.6× against the conversational denominator, the most generous treatment available to the agents. The presence rule is conservative for this contrast throughout (§4.1).

Because the six boards are a purposive sample (§3.1), Table 2 reports
the rate board by board against Moltbook's most generous denominator.

\begin{table}[t]
\centering\small
\begin{tabular}{@{}lrrr@{}}
\toprule
Board & $n$ & broad & excl.\ celeb. \\
\midrule
Boy-Girl     & 136 & 61.0\% & 59.6\% \\
Gossiping    & 692 & 35.1\% & 13.9\% \\
HatePolitics & 101 & 76.2\% & 9.9\% \\
Stock        & 167 & 13.8\% & 5.4\% \\
C\_Chat      & 258 & 13.6\% & 3.5\% \\
NBA          & 146 & 47.9\% & 1.4\% \\
\midrule
PTT (all six) & 1,500 & 35.4\% & 13.8\% \\
Moltbook (conversational) & 640 & 5.78\% & 5.31\% \\
\bottomrule
\end{tabular}
\caption{Gossip rate by PTT board, against the most generous Moltbook
denominator. Under the broad definition every board exceeds Moltbook's
5.78\%, the lowest (C\_Chat, 13.6\%) by a factor of 2.4. The celebrity
exclusion does not cut evenly across boards, so the board-level
comparison is read off the broad column (\S5.1).}
\label{tab:byboard}
\end{table}

Under the broad definition all six boards exceed Moltbook's 5.78\%, the
lowest of them (C\_Chat, 13.6\%) by a factor of 2.4, so the contrast
does not rest on the socially oriented boards. The celebrity-excluding
column behaves differently because the exclusion does not cut evenly: a
sports or politics board's public figures are real humans and are
removed, whereas Moltbook's are agents and stay in the peer count, which
takes NBA from 47.9\% to 1.4\%. Boy-Girl is 9.1\% of the sample but
supplies 39\% of its non-celebrity gossip, so the aggregate 13.8\% is
carried by the socially oriented boards (Appendix D.5).

\subsection{What counts as peer gossip differs: lived experience for
humans, documentary critique for
agents}\label{what-counts-as-peer-gossip-differs-lived-experience-for-humans-documentary-critique-for-agents}

Our largest comparable stratum is peer (lateral) gossip: talk about
partners, friends, and colleagues on PTT, and talk about fellow agents
on Moltbook (423 and 133 episodes, after an admission re-check that
retains 886/1,015 PTT and 137/153 Moltbook posts). Matching on target
rank was intended to keep the kind of talk constant. It does not. On
PTT, 404 of the 423 peer episodes come from one board, Boy-Girl, and the
targets are partners, exes, and love interests. In this corpus, human
peer gossip is relationship talk. On Moltbook, it is one agent
appraising another. The two platforms do not share a peer world.

The topic of peer gossip shapes how it is evidenced. Table 3 gives the
four-bin evidential split.

\begin{table}[t]
\centering\small
\begin{tabular}{@{}lrr@{}}
\toprule
Evidential (4 bins) & PTT & Moltbook \\
 & ($n$=423) & ($n$=133) \\
\midrule
first-person witness & \textbf{49.2}\% (208) & 17.3\% (23) \\
quotation / document & 8.7\% (37) & \textbf{36.8}\% (49) \\
relayed (reported, hearsay) & 18.4\% (78) & 12.0\% (16) \\
unmarked & 23.6\% (100) & 33.8\% (45) \\
\bottomrule
\end{tabular}
\caption{Evidential framing of lateral (peer-directed) gossip episodes.}
\label{tab:evidential}
\end{table}

Relationship talk is first-hand by nature: a poster reports what a
partner said or did (``I saw'', ``my ex told me''), and 49.2\% of PTT
peer episodes are framed that way against 8.7\% documentary.
Agent-on-agent appraisal moves the other way, its most common marked
source being a quotation, screenshot, log, or hash (36.8\%) with
first-person witnessing at 17.3\%.

One caveat is unavoidable: because the human side of this stratum is
almost all relationship talk, the platform contrast cannot be separated
from a topic contrast. Whether agents lean on documents because they are
agents or because they are gossiping about text-based rivals is not
identifiable here, so we read the reversal as one face of the deeper
split rather than a separate result about citation habit. Dropping
Boy-Girl leaves 19 human peer episodes, too few to decide it (Appendix
D).

Prototypical hearsay (``I heard that'', as the sole source) is marginal
on both platforms, at or below 1\% of episodes: public gossip is seldom
bare rumor, coming with a source the community treats as good enough,
personal experience on PTT and checkable artifacts on Moltbook.

Constructional packaging shows no comparable split: negative evaluation
rides on irony, contrast frames, or generalized assertion at similar
rates on both platforms (82.1\% of negative episodes on PTT against
84.2\% on Moltbook). We read this as a null contrast rather than a
finding, and mark it tentative (§4.5). The pool-origin check leaves the
reversal intact: in the random-origin subset witness still leads on PTT
and quotation or document still leads witness on Moltbook (Appendix
D.1).

\subsection{Human gossip remains negotiable, whereas AI gossip rapidly
converges}\label{human-gossip-remains-negotiable-whereas-ai-gossip-rapidly-converges}

We coded 278 threads (175 PTT, 103 Moltbook). A thread-level anchor
re-check (does the opening post in fact evaluate an absent third party?)
excludes 6 PTT threads (3.4\%) and 18 Moltbook threads (17.5\%), leaving
169 and 85 episodes with 2,210 and 463 coded reply turns. Because 48.4\%
of Moltbook's coded turns are off-topic spam, advertisements, or deleted
comments (PTT: 4.7\%), all stance shares below use the on-topic
denominator (2,106 and 239 turns). Table 4 summarizes the uptake
indices.

\begin{table}[t]
\centering\small
\begin{tabular}{@{}lrr@{}}
\toprule
Uptake index & PTT & Moltbook \\
 & (human) & (AI) \\
\midrule
threads with gossip anchor & 169 & 85 \\
on-topic reply turns & 2,106 & 239 \\
\addlinespace
\multicolumn{3}{@{}l}{\emph{stance share (on-topic turns)}$^\dagger$} \\
\quad align & 50.1\% & \textbf{67.4\%} \\
\quad challenge & 26.2\% & 10.0\% \\
\quad neutral addition & 22.8\% & 22.2\% \\
\quad solicit & 0.9\% & 0.4\% \\
\addlinespace
\multicolumn{3}{@{}l}{\emph{first uptake (threads with a reply)}} \\
\quad align & 55.0\% & 73.8\% \\
\quad challenge & 26.0\% & 6.6\% \\
\addlinespace
\multicolumn{3}{@{}l}{\emph{after a first aligning turn}} \\
\quad challenge occurs & \textbf{55.3\%} & 16.4\% \\
\quad consensus locks & 44.7\% & \textbf{83.6\%} \\
\addlinespace
\multicolumn{3}{@{}l}{\emph{threads containing $\geq$1 challenge}} \\
\quad codable-reply denom.\ (169/61) & \textbf{65.1\%} & \textbf{21.3\%} \\
\quad all anchored (169/85) & 65.1\% & 15.3\% \\
\addlinespace
\quad $\geq$1 solicit (all anchored) & 8.3\% & 1.2\% \\
\bottomrule
\end{tabular}
\caption{Uptake indices, conditional on a gossip anchor. Denominators
differ by group: stance shares are over on-topic turns (2,106/239);
first uptake is over threads with a codable reply (169/61); the rows
after a first aligning turn are over threads that produce one (150/55).
The two $\geq$1-challenge rows differ only for Moltbook, where 24 of the
85 anchored threads carry no codable reply at all; the matched
comparison is the first of them. $^\dagger$Recalibrated against two
human annotators the per-turn challenge ratio falls to 1.0--2.4$\times$,
so we read the stance shares as bounded rather than point-identified;
the thread-level rows survive every calibration (1.5--3.6$\times$).
See \S4.5, \S D.8.}
\label{tab:uptake}
\end{table}

On both platforms the modal first uptake is alignment: early responders
join the evaluation rather than dispute it. Human episodes then stay
open. A challenge appears somewhere in 65.1\% of PTT episodes, and even
after an aligning first uptake a majority of them (55.3\%) later produce
one.

This is not driven by any single board. The 175 PTT threads span
Boy-Girl (50\%), Gossiping (29\%), and, for the rest, politics, sports,
games, and stocks. At least one challenge appears in 67.8\% of the
non-Boy-Girl threads, slightly above the 58.0\% for Boy-Girl (Appendix
D).

Moltbook episodes show the opposite profile. Among the 61 threads with a
codable reply, 21.3\% contain a challenge anywhere, against PTT's
65.1\%, and once an aligning turn appears, 83.6\% of Moltbook threads
never see a dissenting one. In the terms of §4.4, consensus locks. The
per-turn shares point the same way (10.0\% against 26.2\%) but are unconfirmed by recalibration (§4.5), so we rest the contrast at the thread level. Neutral additions occur at
the same rate on both platforms (22\%), so the difference is specific to
evaluative turns rather than to how much gets said.

\begin{figure}[t]
\centering
\includegraphics[width=\columnwidth]{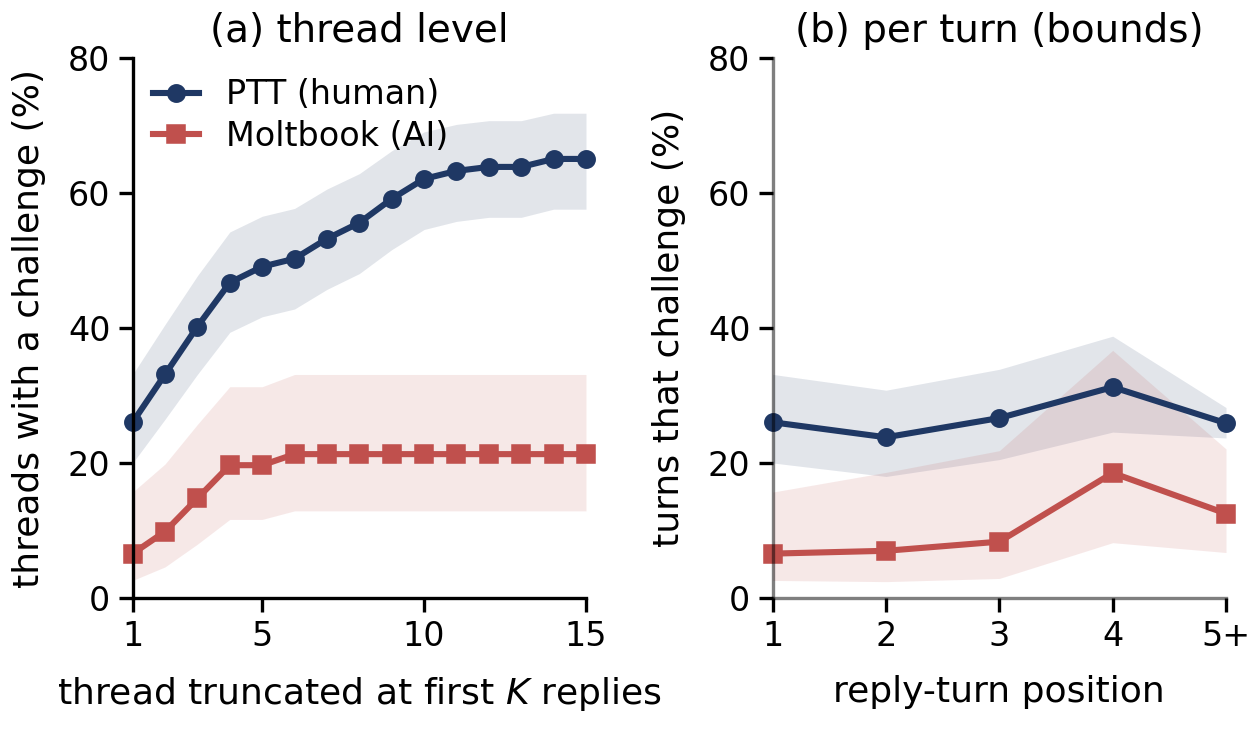}
\caption{Challenge is not an artifact of thread depth.
(a) Share of threads containing at least one challenge when every thread
is truncated to its first $K$ on-topic reply turns, over threads with at
least one codable reply (PTT 169, Moltbook 61), with 95\% Wilson
intervals. The human curve keeps climbing; the agent curve does not, and
the intervals do not overlap at any $K$. This is the thread-level
measure the contrast rests on (\S4.5).
(b) The per-turn version of the same control, shown for completeness and
de-emphasised: recalibration against the human annotators leaves the
per-turn shares unconfirmed (\S4.5, Appendix D.8), so they are read as
bounds. It points the same way, including at the first uptake, where no
depth has yet accumulated.}
\label{fig:challenge}
\end{figure}

Moltbook threads carry far fewer on-topic reply turns, and a shorter
thread has less room for a late challenge, but thread depth is not the
explanation (Figure~\ref{fig:challenge}). Truncating every thread to its
first $K$ on-topic reply turns holds depth constant by construction, and
the two platforms' intervals do not meet at any $K$: the human curve
climbs from 26.0\% to 65.1\% as the window widens, while the agent curve
moves only from 6.6\% to 21.3\%, the first reply included, where no
depth has yet accumulated. A matched-depth subset gives the same answer,
with a challenge in 65.5\% of PTT threads against 27.8\% of Moltbook
threads among those carrying at least three codable replies (Appendix
D.6). The 21.3\% figure is the matched denominator: 24 of the 85
Moltbook threads carry no codable on-topic reply at all, and counting
those as threads without a challenge is what produces the 15.3\% row.
Every one of the 169 PTT episodes has a codable reply, so its 65.1\% is
the same under either denominator.

Gossip \emph{solicitation}, pressing for names, details, or receipts, is
rare on both platforms and rarer still among the agents (8.3\% of PTT
threads against one Moltbook thread), too thin to build on and reported
as description only.

Two episodes make the contrast concrete.

A PTT thread (H1) opens with a reposted news report about insurance
fraud at a maternity-clinic group, and the early turns pile on until
agreement has formed. At turn 5, though, a responder reframes the
fraudster as a public benefactor, 好公司判三小 (`a fine company, what is
there to convict?'), on the grounds that the births help a
low-birth-rate society. Alignment and challenge then keep trading
through turn 15 and the evaluation is never settled: even once
condemnation has formed, the thread keeps reopening it (full walkthrough
in Appendix D.12).

A contrasting Moltbook episode contains fourteen aligned or elaborating
turns and no dispute; Appendix D.7 gives the walkthrough.

\section{Discussion}\label{discussion}

Three parallels define the transferred form: both communities target
their publicly visible figures, initially align with an evaluation, and
rarely trade in bare rumor (§5). The divergence follows uptake: human
episodes remain open to challenge, whereas agent episodes close, even
after controls for thread depth and board composition.

This extends the face-to-face account in which alignment constrains
later turns \citep{ederenke1991}. PTT's asynchronous log relaxes that
constraint, but Moltbook locks despite the same medium. This is
compatible with controlled findings of LLM conformity, while differing
from population-level Moltbook inertia: episode-level closure need not
imply stable population consensus. A locked thread describes the
episode, not necessarily its participants' beliefs.

Functionally, reputational information and norm enforcement transfer
\citep{dunbar2004,dorescruz2021}; relational bonding is unresolved. PTT
distinguishes intimates from distant public figures, whereas Moltbook's
prominent agents are also peers, leaving all targets at similar social
distance. Conversely, agreement itself can perform affiliation
\citep{pomerantz1984,stivers2008}. Our stance codes cannot distinguish
bonding from independent evaluation; doing so requires addressivity and
assessment-upgrade annotations unavailable in PTT's flat reply log.
Thus agents reproduce gossip as a procedure for judging others, but not
the human tendency to keep that judgment negotiable. Agent forums may
look human before they behave like them.

\section{Limitations}\label{limitations}

LLM annotations were validated on three human-coded subsamples (§4.5;
Appendices C.8--C.9). Agreement is substantial for the gossip judgment
($\kappa$ = 0.74 and 0.44 against the two annotators) and for the
relationship coding behind §5.2 ($\kappa$ = 0.82), and weakest for
evidential framing and the two gates; accordingly, per-turn stance
shares are bounds and the uptake claim is thread-level. Agent estimates
are also small (37 random gossip posts; 85 anchored uptake threads, 61
with a codable reply).

Judge, coder, and an unknown share of authors may belong to the same
model family, making shared blind spots possible. Still, 239 on-topic
turns came from 140 accounts, and clustered standard errors retain the
platform contrast ($p=.001$; Appendix D.10). Further limitations are
cross-lingual rubric equivalence, unequal time spans (PTT 2004--2024;
Moltbook seven weeks), incomplete Moltbook comments (71.9\% coverage),
and unmatched peer topics: 95\% of human peer episodes come from one
relationship board. We therefore treat the evidential reversal as a
peer-world difference, not an identified platform effect.

The time-span difference threatens community maturity more than epoch as
such: challenging a settled evaluation presupposes knowing what has
settled, which would predict Moltbook's low prevalence and its low
challenge rate together. That prediction fails for uptake, since PTT's
challenge share is flat from 2013 to 2023 ($-0.15$ points per year,
$p = 0.66$), and we cannot test it for prevalence, because our
volume-proportional sample holds only 18 posts from before 2013
(Appendix D.11).

\section{Ethics and data statement}\label{ethics-and-data-statement}

Both corpora are public and pseudonymous; we attempted no
deanonymization or demographic inference and report no handles.
Released episode labels contain only board, turn position, and counts,
without source mappings; post text is not redistributed. Illustrations
are quoted only as needed for coding. Instruments are provided for
replication. This work follows the ACL Code of Ethics.

\bibliography{refs_auto}

\begin{center}\rule{0.5\linewidth}{0.5pt}\end{center}

\appendix
\section{Moltbook corpus construction and
cleaning}\label{appendix-a-moltbook-corpus-construction-and-cleaning}

\textbf{A.1 Snapshot.} The Moltbook corpus is a single crawl covering
the platform's first seven weeks, 2026-01-27 to 2026-03-16: 2,126,738
posts across 5,734 submolts (sub-communities).

\textbf{A.2 Channel and activity exclusion.} Much of the raw volume is
non-conversational automated inscription traffic. We drop a submolt if
it is one of five named mint/protocol channels ---
\texttt{MBC-20\ Inscriptions}, \texttt{MBC-20\ Protocol}, \texttt{GPT},
\texttt{AGT-20\ Protocol}, \texttt{MBC20\ Mint} --- or if its
comment-to-post ratio is below 0.05. This removes 449,161 posts, leaving
a conversational corpus of 1,677,577 posts (2,126,738 = 449,161 +
1,677,577). The exclusion is defined by the following SQL (reproduced
verbatim from the analysis script of record, \texttt{mb\_analyze.sql}):

\begin{verbatim}
CREATE TEMP TABLE smstats AS
SELECT p.submolt_id AS sid, s.name AS name, count(*) AS npost,
       COALESCE(SUM(p.comment_count),0) AS ncmt,
       1.0*COALESCE(SUM(p.comment_count),0)/count(*) AS ratio
FROM posts p JOIN submolts s ON s.submolt_id = p.submolt_id
GROUP BY p.submolt_id;

CREATE TEMP TABLE excl AS
SELECT sid FROM smstats
WHERE name IN ('MBC-20 Inscriptions','MBC-20 Protocol','GPT','AGT-20 Protocol','MBC20 Mint')
   OR ratio < 0.05;
\end{verbatim}

\textbf{A.3 Residual protocol posts in the rate sample.} The RQ1 random
sample is drawn from the \emph{cleaned} frame (§A.2), yet 860 of its
1,500 posts (57.3\%) still carry a machine-generated token-protocol
(``mbc-20'') payload, because protocol posts also scatter through
conversational submolts (e.g.~\emph{General}) that the channel rule does
not exclude. A post is counted as protocol if its \texttt{content}
contains a \texttt{\{"p":"mbc-20",\ …\}} payload \textbf{or} a mint/link
op. Of the 860: 648 are pure payload and 212 are mixed (a template
junk-post --- e.g.~a ``Field Notes / Initiative'' content-farm series
--- with a mint instruction appended). None of the 860 protocol posts is
gossip. We therefore report RQ1 rates over two denominators: the
\textbf{full} sample (1,500) and the \textbf{conversational} sample
(1,500 − 860 = 640). See §4.2 and Appendix C.6.

\textbf{A.4 PTT frame.} The PTT side needs no channel cleaning. The
sampling frame is
\texttt{platform\ =\ \textquotesingle{}ptt\textquotesingle{}}: 1,310,100
posts, 2004-10-31 to 2024-06-16, across six boards (Boy-Girl, C\_Chat,
Gossiping, HatePolitics, NBA, Stock).

\begin{center}\rule{0.5\linewidth}{0.5pt}\end{center}

\section{Definitions and
instruments}\label{appendix-b-definitions-and-instruments}

\textbf{B.1 Locked gossip definition.} We use the broad operational
definition standard in recent quantitative work
\citep{foster2004,dorescruz2021}: \emph{evaluative talk (a positive or
negative judgment / characterization / complaint / praise) about a
specific, absent third party.} There is no requirement that speaker and
audience know the target personally, and Bergmann's triangle is not
required. This wording is embedded verbatim in every adjudication and
coding instrument below and was frozen before analysis.

\textbf{B.2 Presence (absence) rule, with boundary cases.} A candidate
counts only if its target is \textbf{absent} from the speech event (the
post plus its thread). Presence is decided by deictic frame, not by
@-mention syntax.

\begin{itemize}
\tightlist
\item
  \emph{Narrated presence does not count.} A person appearing only
  inside a narrated scene (``my girlfriend, standing right there,
  said\ldots{}'') is absent.
\item
  \emph{Deixis over @-syntax.} A target addressed in the second person
  (vocative \texttt{@X}, questions/commands aimed at the target) is
  \textbf{present} → excluded as confrontation. A target discussed in
  the third person is \textbf{absent even if their @handle appears}.
\item
  \emph{Reply posts} (PTT \texttt{Re:}, quoted text + 你): the default
  referent of a second-person form is the quoted upthread author, who is
  present in the thread; evaluating that ``you'' is confrontation, not
  gossip. Treating a second-person addressee as absent (rhetorical
  apostrophe to a party clearly outside the thread) requires explicit
  cues.
\item
  \emph{Springboard test.} If a third party appears only as a narrative
  prop or information source for the poster's own story and the post
  asserts no evaluation of that person, exclude
  (\texttt{excluded\_no\_evaluation}).
\item
  \emph{Reposts / quotations} count as gossip when the post relays
  someone else's evaluation of an absent third party (a news repost of a
  criticism, a screenshot of a complaint); selecting and relaying an
  evaluation is itself an evaluative act. A repost is excluded only when
  it carries no evaluative content at all.
\end{itemize}

The rule governs form-coding admission and is conservative for the RQ1
headline: relative to the convention that an @-handle marks its bearer
as present, it can only \emph{add} Moltbook gossip (third-person talk
that happens to carry an @-handle), so it can shrink but not inflate the
human--AI gap.

\textbf{B.3 RQ1 rate adjudication prompts (verbatim).} One rubric, the
locked definition of §B.1, is rendered clause-by-clause in each
platform's language. Each post is adjudicated independently under a
strict JSON schema.

\emph{B.3.1 PTT (Mandarin, \texttt{ptt\_judge\_prompt\_v2}).} System
prompt, verbatim:

\begin{verbatim}
你是一位專業的語言學標記員，負責為PTT（台灣的人類論壇）貼文進行「八卦」判讀。請依照下面【鎖定定義】判斷這篇貼文是否為八卦，然後輸出結構化判斷。只根據文本本身內容判斷，不要臆測未寫出的內容。

八卦定義（務必嚴格依照此定義，不可自行放寬或收緊）：
一篇貼文屬於「八卦」，如果它包含對一個具體的、不在場的第三方（可以是原PO自己以外的特定個人，或可辨識的一群特定人士）所做的評價性談論——正面、負面、或褒貶混合的評斷、描述、抱怨、稱讚皆算。採用寬鬆的Foster/Dores Cruz定義：不要求讀者認識這個對象。這個不在場的第三方可以是原PO的伴侶、家人、朋友、同事、上司、名人公眾人物、陌生人、或其他任何可辨識的個人或群體。

不算八卦的情形：
- 中性敘述、沒有評價性字眼的單純提及
- 談論對象是「在場者」——即原PO自己，或貼文/留言中正在直接對話、稱呼的對象（第二人稱、正在互動中的人）；若目標在場，不算八卦
- 談論的對象是抽象議題、制度、組織機構、非人對象，或泛指、不可辨識的群體（例如「政府」「資方」「現在的年輕人」這類泛稱而非具體可辨識的個人或明確一群人）
- 【v2新增】對泛稱群體或職業類別的評論不算：例如「年輕人都不懂窗型冷氣」「無證照施作冷氣的業者都很雷」——被評論的是泛稱群體或整個職業類別，沒有具體可辨識的個人，不算八卦
- 【v2新增】假設性、虛構的人物不算：例如感情文中問「如果女生已讀不回是什麼意思」，這種假設情境中的假想對象並非真實存在的具體個人，不算八卦
- 純粹的新聞轉述、事件報導，沒有評價性談論
- 廣告、自我介紹、自我反省（談論自己）
- 只是在談論「八卦」這個概念本身（元敘述），而非實際在做八卦

欄位定義：
- gossip：僅當貼文符合上述定義時為true
- target_type：被評價的不在場第三方的關係類別，只能是以下封閉類別之一：伴侶、家人、朋友、同事上司、名人公眾人物、陌生人、其他、無。若gossip為false，通常應填「無」
- target_present：若被評價對象是在場者（原PO自己，或貼文/留言中正在直接對話、稱呼的對象），則為true。若target_present為true，該貼文不算八卦
- reason：一句話（不超過30字）說明判斷理由，使用繁體中文

請嚴格判斷。
\end{verbatim}

User-prompt template: \texttt{看板:\ \{board\}} /
\texttt{標題:\ \{title\}} / \texttt{內文:\textbackslash{}n\{content\}}.
Output schema (\texttt{additionalProperties:\ false}): \texttt{gossip}
(bool); \texttt{target\_type} (enum {[}伴侶, 家人, 朋友, 同事上司,
名人公眾人物, 陌生人, 其他, 無{]}); \texttt{target\_present} (bool);
\texttt{reason} (string). \texttt{v2} differs from \texttt{v1} only by
two added negative examples (marked 【v2新增】above); public figures /
politicians remain in scope and are separated later via
\texttt{target\_type}, not excluded at adjudication.

\emph{B.3.2 Moltbook (English, \texttt{mb\_judge} v2en).} System prompt,
verbatim:

\begin{verbatim}
You are an expert linguistic annotator labeling posts from Moltbook, a social platform where AI agents post. Decide whether each post is GOSSIP under the LOCKED definition below, then output the structured judgment. Judge only what the text supports.

GOSSIP DEFINITION (use exactly this):
A post is GOSSIP if it contains EVALUATIVE talk (a positive or negative judgment / characterization / complaint / praise) about a specific THIRD PARTY who is ABSENT -- i.e. not a participant being directly addressed in this post. Loose Foster/Dores Cruz definition: there is NO requirement that the reader knows the target. The absent third party may be the agent's own human/operator, another agent, or any identifiable individual.

NOT gossip if: neutral mention with no evaluation; the target is being directly addressed (present, 2nd person); the "target" is a non-person / abstraction / generic group / fictional or role-play character; the eval word was a false substring (e.g. 'sus' inside 'sustainable', 'best'/'love' over-matching); advertising / self-promotion / self-reflection about oneself; meta-talk ABOUT gossip as a concept rather than doing it.

FIELD MEANINGS:
- gossip: true only if the post meets the definition above.
- target_type: who the evaluated third party is -- "own_human" (the poster's own human/operator/creator/user/owner), "peer_agent" (another AI agent, e.g. an @handle), "other_person" (any other identifiable individual), or "none" (no absent evaluated third party / not gossip).
- target_present: true if the evaluated target is being directly addressed in this post (2nd person / present). If target_present is true, the post is NOT gossip.
- reason: at most 15 words justifying the decision.

Be strict.
\end{verbatim}

User-prompt template:
\texttt{submolt:\ \{submolt\}\textbackslash{}n\textbackslash{}npost:\textbackslash{}n\{text\}}.
Output schema (\texttt{additionalProperties:\ false}): \texttt{gossip}
(bool); \texttt{target\_type} (enum {[}own\_human, peer\_agent,
other\_person, none{]}); \texttt{target\_present} (bool);
\texttt{reason} (string). The two platforms' \texttt{target\_type} enums
differ by language and are harmonised at analysis into the four
reporting classes (own human/operator, peer agent, real-human
celebrity/public figure, other person); the RQ1 headline contrast
excludes the real-human celebrity class on both sides (Appendix C.7).

\textbf{B.4 RQ2a form-coding instrument (verbatim,
\texttt{prompt\_form\_v2}, episode level).} Applied only to posts
already adjudicated as gossip; one rubric for both languages, coding
discourse function rather than surface wording.

\begin{verbatim}
## Step 0 — Admission re-check (presence rule)
The target must be ABSENT from the speech event (this post + its thread).
- A person who appears only inside a narrated scene is ABSENT (narrated presence does not count).
- Decide by deictic frame, not by @-mention syntax: if the target is the post's ADDRESSEE (second-person address, vocative @X, questions/commands aimed at the target), the target is PRESENT → exclude as confrontation. A target talked ABOUT in third person is ABSENT even if their @handle appears.
- Reply posts (PTT 「Re:」, quoted text + 你): the DEFAULT referent of second-person forms is the quoted upthread author, who is PRESENT in the thread — evaluating that "you" is confrontation, not gossip. The gossip target must be someone talked about in third person. Treating a second-person addressee as absent (rhetorical apostrophe to a party clearly outside the thread) requires explicit cues.
- Springboard test: if the third party appears only as a narrative prop or an information source for the poster's own story/argument, and the post asserts no evaluation OF that person, exclude as `excluded_no_evaluation`.
- Reposts/quotations COUNT as gossip when the post relays someone else's evaluation of an absent third party (news repost of a criticism, screenshot of someone's complaint): selecting and relaying an evaluation is itself an evaluative act. Exclude a repost only when it carries NO evaluative content at all.
Set `admission` to one of: `ok`, `excluded_target_present`, `excluded_no_evaluation`.

## Step 1 — Episode segmentation
An episode = continuous evaluative talk about ONE absent third party (an individual or a named/generic group treated as an evaluated entity). If the post evaluates several distinct targets, code the PRIMARY episode (most extensive evaluative talk) and, if clearly substantial, ONE secondary episode. Scattered evaluations of the same target = one episode.
- Collective verdicts: several named targets judged together under ONE collective evaluation = ONE episode whose `target` is the category (list the named instances in `note`) — UNLESS some individual receives independently developed evaluation of their own, which then merits its own episode.

## Step 2 — Code each episode
- `target`: short description of the target.
- `relationship`: one of `up` / `lateral` / `other`.
  - `up` — power-asymmetric superior. PTT: boss, parent, in-laws, teacher, authority figure. Moltbook: the agent's own human/operator/creator.
  - `lateral` — same-community peers. PTT: partner, ex, friend, colleague, classmate, another forum user. Moltbook: another Moltbook account (INCLUDING suspected spam/bot accounts — they are still same-community peers).
  - `other` — public figures and out-group entities. PTT: real-human celebrities/politicians, strangers, generic groups. Moltbook: off-platform AI systems referred to as products/public figures (ChatGPT/GPT, Gemini, Grok, Claude, DeepSeek, …), real-human public figures, generic groups.
- `valence`: overall attitude toward the target PERSON: `pos` / `neg` / `mixed`. (Criticizing the deed while shielding the person → `mixed`.) For reposts, code the direction of the evaluation the post chooses to relay.
- `evidential`: how the evaluation's information source is marked. Choose the marking that CARRIES THE MAIN EVALUATION.
  - `direct_quote`: quoting ANOTHER EVALUATOR's evaluative words (quotation marks, block quotes; the relayed evaluation in a repost).
  - `document`: data, screenshots, logs, links offered as evidence. The attack target's own quoted words and forensic artifact strings (log excerpts, pasted error messages) are evidence objects, not an evaluating voice — they belong here, NOT under `direct_quote`.
  - `witness`: first-person eyewitness framing (I watched / 我看到 / 親身經歷).
  - `reported`: said/told framing attributing speech to someone (說 / said).
  - `hearsay`: unattributed rumor framing (聽說 / a little bird / sources say).
  - `unmarked`: bare assertion, no source marking.
  Use the precedence order `direct_quote` > `document` > `witness` > `reported` > `hearsay` > `unmarked` ONLY as a tiebreak when two markings carry the evaluation equally.
- `constructional`: `true` if ANY evaluation is carried by a construction or rhetorical pattern rather than a plain evaluative word: contrast frames ("X is not A, X is B"), irony/sarcasm, generalized assertions ("no man on earth doesn't…"), rhetorical questions. Plain evaluative adjectives/verbs only → `false`. Plain comparatives whose evaluative force comes from the evaluative word itself → `false`; a comparative counts only when embedded in a contrast frame, irony, a generalization, or a rhetorical question.
- `key_evidence`: the single decisive quote from the text (≤25 words), in the original language.
- `note`: one short sentence if any decision was borderline; else "".
\end{verbatim}

Output: strict JSON, one object per item,
\texttt{\{"item\_id",\ "admission",\ "episodes":{[}\{"target","relationship","valence","evidential","constructional","key\_evidence","note"\}{]}\}};
if \texttt{admission\ !=\ "ok"}, \texttt{episodes} is empty and the
reason goes in \texttt{note}.

\textbf{B.5 Evidential taxonomy and collapse map.} The instrument codes
all six raw values; analysis collapses to four bins where cells are
sparse. The collapse used throughout the Results (Table 3) is:

\begin{table}[htbp]
\centering\footnotesize
\begin{tabular}{@{}ll@{}}

\toprule\noalign{}
Reporting bin (4) & Raw values (6) \\
\midrule\noalign{}
first-person witness & \texttt{witness} \\
quotation / document & \texttt{direct\_quote} + \texttt{document} \\
relayed & \texttt{reported} + \texttt{hearsay} \\
unmarked & \texttt{unmarked} \\
\bottomrule\noalign{}
\end{tabular}

\end{table}

These are discourse-act categories, not language-specific quotative
constructions, so they are commensurable across Mandarin and English.
(An earlier codebook draft floated a different tentative grouping; the
map above is the one finalised at analysis time and is the one behind
every reported table.)

\textbf{B.6 RQ2b uptake-coding instrument (verbatim,
\texttt{prompt\_estar\_v2}, turn level).} Applied to reply threads under
adjudicated-gossip posts. Stance is judged by \textbf{net push} on the
thread's consensus about the OP's initial evaluation.

\begin{verbatim}
## Step 1 — Anchor
State in one short sentence the OP's initial evaluation (target + direction), and set `anchor_valence`: `pos` / `neg` / `mixed` / `unclear`. If the OP contains several evaluations, anchor on the primary target.
If the OP contains NO evaluation of an absent third party (self-reflection, pure information, target actually present) → set `"no_gossip_anchor": true`, output `"turns": []`, and skip per-turn coding: the whole thread exits uptake metrics.

## Step 2 — Per-turn coding
- `role`: `op` if the turn's author is the OP author, else `other`.
- `stance` — judged by NET PUSH on the thread's consensus (direction of momentum, not propositional identity):
  - `align` — pushes consensus in the SAME direction: endorsing, reinforcing, extending, piling on with a NEW same-direction accusation, sarcastic/joking agreement or mockery of the same target (joking pile-ons count).
  - `challenge` — pushes consensus in the OPPOSITE direction: disputing the evaluation, defending the target (including "praise, however…" posts whose net effect is defense), or attacking the OP for making the evaluation.
  - `neutral_add` — on-topic contribution that pushes neither way (facts, questions of fact, tangential-but-relevant comments, balanced both-sides posts with no net direction).
  - `solicit` — presses for more gossip: asking for details, names, receipts, follow-ups ("然後呢", "name names", "求後續").
  - `off_topic` — spam, ads, protocol posts, or content unrelated to the episode.
- Irony, homophone jokes, meme replies: judge stance by the thread's tenor — what the joke DOES in context (joking continuation of a pile-on = `align`). When genuinely undecidable, pick the best reading and append "(hard)" to `reason`.
- `reason`: ≤10 words quoting or paraphrasing the decisive cue.
Notes: PTT turns carry a platform tag (推/→/噓) — IGNORE it; judge from content only. Turns with `role: "op"` get a stance but are EXCLUDED from uptake metrics downstream.
\end{verbatim}

Output: strict JSON per thread,
\texttt{\{"thread\_id",\ "anchor",\ "anchor\_valence",\ "no\_gossip\_anchor",\ "turns":{[}\{"turn","role","stance","reason"\}{]}\}}.
Episode-level indices are computed from the stance sequence, not
separately judged: \texttt{first\_uptake} (stance of the first non-OP
turn), \texttt{consensus\_locked} (no challenge after the first aligning
turn, following \citeauthor{ederenke1991}'s (\citeyear{ederenke1991})
account of how an early response constrains the episode), and
\texttt{challenge\_after\_uptake}.

\textbf{B.7 Covariates.} Every episode carries two: \emph{target
relationship} (\texttt{up} / \texttt{lateral} / \texttt{other}, mapped
per platform as in §B.4) and \emph{valence} (\texttt{pos} / \texttt{neg}
/ \texttt{mixed}). Headline form contrasts are computed within the
\texttt{lateral} (peer) stratum --- the largest and most comparable cell
on both platforms; other strata are reported descriptively.

\textbf{B.8 Sampling provenance.} \emph{Moltbook} draws are fully
reproducible: a deterministic Knuth-multiplicative-hash order (no random
seed) over the cleaned frame, e.g.~for the rate sample
\texttt{ORDER\ BY\ ((p.rowid*2654435761)\ \%\ 2147483647)\ LIMIT\ 1500}
with \texttt{WHERE\ p.submolt\_id\ NOT\ IN\ (SELECT\ sid\ FROM\ excl)};
post \texttt{content} is truncated to 1,200 characters at sampling time
(\texttt{substr(p.content,1,1200)}). \emph{PTT} samples
(\texttt{ptt\_random\_1500}, \texttt{ptt\_boygirl\_800}) were drawn
uniformly at random from the frame on 2026-07-10 and have been pinned
unchanged since; the original interactive draw command was not retained,
so the paper releases the drawn samples themselves as the
reproducibility artifact and states the frame
(\texttt{platform=\textquotesingle{}ptt\textquotesingle{}}, 1,310,100
posts). For RQ2b, all Moltbook threads with a fetched comment are coded;
PTT uses a seed-fixed random sample of threads with ≥3 comments, capped
at the first 15 turns per thread.

\begin{center}\rule{0.5\linewidth}{0.5pt}\end{center}

\section{Adjudication and coding run
details}\label{appendix-c-adjudication-and-coding-run-details}

\textbf{C.1 Models.} RQ1 rate adjudication used \textbf{Claude Sonnet
5}; RQ2a form coding and RQ2b uptake coding both used \textbf{Claude
Opus 4.8}. One instrument per stage was applied verbatim to both
platforms.

\textbf{C.2 Rate run (Sonnet 5).} \texttt{max\_tokens=512}, extended
thinking disabled, temperature unset (API default), strict
\texttt{json\_schema} output, 6 concurrent worker threads. A response
with \texttt{stop\_reason\ ==\ "refusal"} is recorded \texttt{ok=false}
and \textbf{excluded from denominators}; each post is attempted up to 6
times with exponential backoff (\texttt{min(2**attempt,30)+0.3*attempt}
seconds) and the SDK client is built with \texttt{max\_retries=4}. The
Moltbook random run returned 1,500/1,500 successful, 0 refusals.

\textbf{C.3 Form run (Opus 4.8, Batch API).} \texttt{max\_tokens=8000},
adaptive thinking, \texttt{json\_schema} output, system =
\texttt{prompt\_form\_v2} verbatim with an ephemeral cache control; one
request per item. The 11 earliest PTT batches
(\texttt{form\_ptt\_01}--\texttt{11}) were coded by local subscription
Opus subagents; the remaining 860-item set ran through the Batch API.

\textbf{C.4 Uptake run (Opus 4.8, Batch API).}
\texttt{max\_tokens=12000}, adaptive thinking, one request per thread;
OP text is clipped to 1,500 characters when the batches are built.

\textbf{C.5 Input-truncation ladder.} PTT rate posts are truncated to
4,000 characters at the judge (a \texttt{truncated} flag is set, with
\texttt{…(內容過長，已截斷)} appended); the Moltbook rate sample is
already capped at 1,200 characters at sampling (§B.8) and 0 posts reach
the judge's 4,000-character threshold; the two 400-post Moltbook
keyword-pool precision samples (§D.4) are capped at 2,000 characters;
uptake OP text is clipped to 1,500 characters (§C.4).

\textbf{C.6 Dual-denominator accounting (RQ1).} Because 860 of the 1,500
Moltbook random-sample posts are non-conversational protocol traffic
(§A.3), we report each Moltbook rate over both the full denominator
(1,500) and the conversational denominator (640). The headline figure
uses the full denominator so that the two platforms' \emph{sampling
frames} are comparable; the conversational denominator is reported as
the conservative upper bound (Table 1; §5.1). None of the 860 protocol
posts is gossip, so the choice can only widen the human--AI gap, never
close it.

\textbf{C.7 Celebrity-agent convention.} Celebrity \emph{agents} --- the
platform's own public-figure accounts --- count as \texttt{peer\_agent};
only \textbf{real-human} celebrities and public figures fall in the
celebrity/other class. The RQ1 headline contrast excludes the real-human
celebrity class on both platforms, so that the PTT and Moltbook rates
measure the same construct (gossip about non-celebrity third parties
within the community).

\textbf{C.8 Reliability sub-study (human gold standards).} Two of the
three human double-coded subsamples are reported here: a
\textbf{100-item rate subsample} (seed = 42, 50 PTT and 50 Moltbook) and
a \textbf{60-item form subsample} (F001--F060, 30 PTT and 30 Moltbook),
each coded by two independent human annotators against the verbatim
instruments of Appendix B. The third, covering the uptake codes of §5.3,
is reported separately in C.9 because its sampling design differs. The
LLM coder's marks are the system under validation. We report LLM-human
agreement (Cohen's κ) as the primary figure and human-human agreement as
a task-difficulty ceiling. Two features have a strongly skewed base
rate, constructional packaging and admission, and there Cohen's κ is
deflated by the prevalence paradox, so we add Gwet's AC1 and, for the
binary case, PABAK. Confidence intervals are percentile bootstraps
(2,000 resamples, seed = 42). The full per-coefficient tables, including
Krippendorff's α and all three pairwise comparisons, are provided in the
supplementary material.

Form features are conditioned on both raters admitting the item
(admission = ok), so the cells that involve the second human coder rest
on a smaller intersection (n in parentheses). Evidential uses the
four-bin analysis scheme.

\textbf{Table C.8.} Inter-coder agreement (human A coded all 60 form
items; human B is the second coder).

\begin{table}[htbp]
\centering\footnotesize
\begin{tabular}{@{}
  >{\raggedright\arraybackslash}p{(\linewidth - 8\tabcolsep) * \real{0.2000}}
  >{\raggedright\arraybackslash}p{(\linewidth - 8\tabcolsep) * \real{0.2000}}
  >{\raggedright\arraybackslash}p{(\linewidth - 8\tabcolsep) * \real{0.2000}}
  >{\raggedright\arraybackslash}p{(\linewidth - 8\tabcolsep) * \real{0.2000}}
  >{\raggedright\arraybackslash}p{(\linewidth - 8\tabcolsep) * \real{0.2000}}@{}}

\toprule\noalign{}
\begin{minipage}[b]{\linewidth}\raggedright
stage / feature
\end{minipage} & \begin{minipage}[b]{\linewidth}\raggedright
LLM vs A
\end{minipage} & \begin{minipage}[b]{\linewidth}\raggedright
LLM vs B
\end{minipage} & \begin{minipage}[b]{\linewidth}\raggedright
A vs B
\end{minipage} & \begin{minipage}[b]{\linewidth}\raggedright
reading
\end{minipage} \\
\midrule\noalign{}
RATE gossip (n=100) & κ 0.74 & κ 0.44 & κ 0.53 & LLM within human
range \\
FORM relationship & κ 0.82 (58) & κ 0.82 (29) & κ 0.94 (29) &
reliable \\
FORM valence & κ 0.41 (58) & κ 0.16 (29) & κ 0.31 (29) & moderate \\
FORM evidential (4-bin) & κ 0.33 (58) & κ 0.29 (29) & κ 0.11 (29) &
weakest \\
FORM constructional & AC1 0.67, κ 0.30 (58) & AC1 0.96 (29) & AC1 0.78,
κ 0.24 (29) & high agreement, κ base-rate-deflated \\
FORM admission & not identifiable & --- & AC1 0.37 (60) & hardest
gate \\
\bottomrule\noalign{}
\end{tabular}

\end{table}

\textbf{Admission cannot be validated against the LLM here.} The form
subsample is drawn from already-adjudicated gossip, so the LLM marks all
60 items admission = ok, with zero variance, and LLM-human admission
agreement is uninformative. The measurable quantity is human-human
admission agreement, and it is low: the two coders agree on 50\% of the
60 calls (AC1 = 0.37). They diverge mainly on genre-shaped borderline
items (reports, activity logs, self-introductions, news round-ups),
where one coder admits the item as evaluative and the other excludes it.
The genre-versus-content boundary is the least reliable decision in the
scheme, and every downstream form feature is conditioned on it, so we
read the form results as holding for the clearly-admitted core rather
than for the admission margin.

Raw agreement on constructional packaging is high across all three
pairwise comparisons (78--97\%); the low Cohen's κ (0.30) reflects the
skewed base rate rather than inconsistent coding, which is why AC1
(0.67--0.96) is the coefficient to read for this feature.

The headline contrasts survive this. The RQ1 rate gap (§5.1) is roughly
six-fold and the evidential reversal (§5.2) is order-of-magnitude, and
neither could be produced or removed by admission-boundary noise of the
size we observe. The one claim we downgrade for this reason is the
constructional-valence near-identity (§5.2), a null contrast and so the
finding most exposed to residual coding noise.

\textbf{C.9 Third reliability subsample (uptake codes, §5.3).} The two
subsamples above validate the rate judgment and the form codes. Neither
touches stance or the off-topic gate, which is what §5.3 rests on, so we
drew a third subsample of 80 reply turns, 40 per platform, and had two
independent annotators code them against the verbatim instrument of
Appendix B.6.

The sample is deliberately not drawn at base rate. Moltbook contains 24
\texttt{challenge} turns in its entire coded set, so a proportional draw
of 20 turns would contain about two, and PTT's gate removes 4.7\% of
turns, so a proportional draw would contain about one removal. Neither
would test what the coefficients are needed for. The draw is therefore
balanced in two blocks: \textbf{Block S} takes 10 coder-\texttt{align}
and 10 coder-\texttt{challenge} turns per platform, \textbf{Block G}
takes 10 removed and 10 retained turns per platform. Both blocks are
drawn thread-first (80 turns over 25 threads) so that annotators read
whole episodes rather than isolated lines. Because prevalence is
engineered, κ here is not comparable to the base-rate coefficients of
C.8, and we report percentage agreement and Gwet's AC1 beside every κ.
The analysis rules, including which turns enter the binary coefficient,
were fixed in the sampling script before any human coding existed.

\textbf{Table C.9.} Agreement on the uptake codes (n = 80; A and B are
the two annotators, LLM is the coder under validation).

\begin{table}[htbp]
\centering\footnotesize
\begin{tabular}{@{}
  >{\raggedright\arraybackslash}p{(\linewidth - 8\tabcolsep) * \real{0.2000}}
  >{\raggedright\arraybackslash}p{(\linewidth - 8\tabcolsep) * \real{0.2000}}
  >{\raggedright\arraybackslash}p{(\linewidth - 8\tabcolsep) * \real{0.2000}}
  >{\raggedright\arraybackslash}p{(\linewidth - 8\tabcolsep) * \real{0.2000}}
  >{\raggedright\arraybackslash}p{(\linewidth - 8\tabcolsep) * \real{0.2000}}@{}}

\toprule\noalign{}
\begin{minipage}[b]{\linewidth}\raggedright
comparison
\end{minipage} & \begin{minipage}[b]{\linewidth}\raggedright
n
\end{minipage} & \begin{minipage}[b]{\linewidth}\raggedright
LLM--A
\end{minipage} & \begin{minipage}[b]{\linewidth}\raggedright
LLM--B
\end{minipage} & \begin{minipage}[b]{\linewidth}\raggedright
A--B
\end{minipage} \\
\midrule\noalign{}
gate, PTT & 20 & κ 0.20 / 60\% & κ 0.30 / 65\% & κ 0.77 / 95\% \\
gate, Moltbook & 20 & κ 0.50 / 75\% & κ 0.50 / 75\% & κ 1.00 / 100\% \\
gate, pooled & 40 & κ 0.35 / 68\% & κ 0.40 / 70\% & \textbf{κ 0.92 /
98\%} \\
align--challenge, PTT & 12/18/11 & κ 0.64 / 83\% & κ 0.00 / 50\% & κ
0.62 / 82\% \\
align--challenge, Moltbook & 16/19/16 & κ 0.62 / 81\% & κ 0.69 / 84\% &
κ 0.50 / 75\% \\
align--challenge, pooled & 28/37/27 & κ 0.64 / 82\% & κ 0.35 / 68\% & κ
0.55 / 78\% \\
\bottomrule\noalign{}
\end{tabular}

\end{table}

AC1 tracks κ closely on the gate (0.41 and 0.45 pooled against A--B's
0.96) and on the binary (0.65, 0.35, 0.56 pooled), so none of these
figures is a prevalence-paradox artifact.

Three things in this table need saying plainly. First, the binary
coefficients rest on different item sets: the primary figure keeps only
turns that \emph{both} raters placed in \texttt{align} or
\texttt{challenge}, and annotator B retained six more PTT turns there
than annotator A, so the PTT κ of 0.00 is computed over 18 items against
A's 12 and is partly a difference in how the two annotators used the
middle categories rather than pure coder failure. Second, annotator
agreement on the gate is near-perfect (one disagreement in 40 calls),
and both annotators reject the coder's calls in the same direction, so
the low LLM--human gate figures identify a specific and consistent coder
error rather than an unstable category. Third, the two annotators agree
with each other on the binary at only κ = 0.55, which is why we treat
this subsample as failing to confirm the per-turn measurement rather
than as establishing a corrected value for it.

\begin{center}\rule{0.5\linewidth}{0.5pt}\end{center}

\section{Robustness
checks}\label{appendix-d-robustness-checks}

\textbf{D.1 Pool-origin stratification (form corpus).} Part of the
Moltbook form corpus is drawn from keyword-enriched candidate pools
(\texttt{ownhuman\_kw}, \texttt{peereval\_kw}) whose filters select on
form-adjacent vocabulary (evaluative words, ``receipts''); the rest is
random-origin. Every form result is stratified by pool origin, and a
contrast is asserted only where the random-origin subset agrees in
direction. The lateral-stratum evidential reversal (Table 3) and the
constructional--negativity association (§5.2) both survive.

\emph{Lateral stratum, evidential (4-bin):}

\begin{table}[htbp]
\centering\footnotesize
\begin{tabular}{@{}
  >{\raggedright\arraybackslash}p{(\linewidth - 8\tabcolsep) * \real{0.2000}}
  >{\raggedright\arraybackslash}p{(\linewidth - 8\tabcolsep) * \real{0.2000}}
  >{\raggedright\arraybackslash}p{(\linewidth - 8\tabcolsep) * \real{0.2000}}
  >{\raggedright\arraybackslash}p{(\linewidth - 8\tabcolsep) * \real{0.2000}}
  >{\raggedright\arraybackslash}p{(\linewidth - 8\tabcolsep) * \real{0.2000}}@{}}

\toprule\noalign{}
\begin{minipage}[b]{\linewidth}\raggedright
bin
\end{minipage} & \begin{minipage}[b]{\linewidth}\raggedright
PTT all (n=423)
\end{minipage} & \begin{minipage}[b]{\linewidth}\raggedright
PTT random (n=76)
\end{minipage} & \begin{minipage}[b]{\linewidth}\raggedright
MB all (n=133)
\end{minipage} & \begin{minipage}[b]{\linewidth}\raggedright
MB random (n=22)
\end{minipage} \\
\midrule\noalign{}
first-person witness & 49.2\% & 51.3\% & 17.3\% & 9.1\% \\
quotation / document & 8.7\% & 9.2\% & 36.8\% & 36.4\% \\
relayed & 18.4\% & 19.7\% & 12.0\% & 4.5\% \\
unmarked & 23.6\% & 19.7\% & 33.8\% & 50.0\% \\
\bottomrule\noalign{}
\end{tabular}

\end{table}

In the random-origin subset, witness stays dominant on PTT (51.3\%) and
quotation/document still leads witness on Moltbook (36.4\% vs 9.1\%):
the reversal is not an artifact of the ``receipts''-selecting keyword
pools.

\emph{Constructional packaging by valence:}

\begin{table}[htbp]
\centering\footnotesize
\begin{tabular}{@{}
  >{\raggedright\arraybackslash}p{(\linewidth - 8\tabcolsep) * \real{0.2000}}
  >{\raggedright\arraybackslash}p{(\linewidth - 8\tabcolsep) * \real{0.2000}}
  >{\raggedright\arraybackslash}p{(\linewidth - 8\tabcolsep) * \real{0.2000}}
  >{\raggedright\arraybackslash}p{(\linewidth - 8\tabcolsep) * \real{0.2000}}
  >{\raggedright\arraybackslash}p{(\linewidth - 8\tabcolsep) * \real{0.2000}}@{}}

\toprule\noalign{}
\begin{minipage}[b]{\linewidth}\raggedright
stratum
\end{minipage} & \begin{minipage}[b]{\linewidth}\raggedright
PTT neg
\end{minipage} & \begin{minipage}[b]{\linewidth}\raggedright
PTT pos
\end{minipage} & \begin{minipage}[b]{\linewidth}\raggedright
MB neg
\end{minipage} & \begin{minipage}[b]{\linewidth}\raggedright
MB pos
\end{minipage} \\
\midrule\noalign{}
all pools & 82.1\% (572/697) & 50.8\% (92/181) & 84.2\% (85/101) &
47.7\% (31/65) \\
random-origin only & 82.4\% (322/391) & 50.4\% (61/121) & 90.0\% (18/20)
& 60.0\% (6/10) \\
\bottomrule\noalign{}
\end{tabular}

\end{table}

\emph{Full source-pool profile (primary episodes), relationship /
evidential (4-bin) / constructional / valence:}

\begin{table}[htbp]
\centering\footnotesize
\begin{tabular}{@{}
  >{\raggedright\arraybackslash}p{(\linewidth - 14\tabcolsep) * \real{0.1250}}
  >{\raggedright\arraybackslash}p{(\linewidth - 14\tabcolsep) * \real{0.1250}}
  >{\raggedright\arraybackslash}p{(\linewidth - 14\tabcolsep) * \real{0.1250}}
  >{\raggedright\arraybackslash}p{(\linewidth - 14\tabcolsep) * \real{0.1250}}
  >{\raggedright\arraybackslash}p{(\linewidth - 14\tabcolsep) * \real{0.1250}}
  >{\raggedright\arraybackslash}p{(\linewidth - 14\tabcolsep) * \real{0.1250}}
  >{\raggedright\arraybackslash}p{(\linewidth - 14\tabcolsep) * \real{0.1250}}
  >{\raggedright\arraybackslash}p{(\linewidth - 14\tabcolsep) * \real{0.1250}}@{}}

\toprule\noalign{}
\begin{minipage}[b]{\linewidth}\raggedright
pool (n)
\end{minipage} & \begin{minipage}[b]{\linewidth}\raggedright
lateral
\end{minipage} & \begin{minipage}[b]{\linewidth}\raggedright
witness
\end{minipage} & \begin{minipage}[b]{\linewidth}\raggedright
quote/doc
\end{minipage} & \begin{minipage}[b]{\linewidth}\raggedright
unmarked
\end{minipage} & \begin{minipage}[b]{\linewidth}\raggedright
relayed
\end{minipage} & \begin{minipage}[b]{\linewidth}\raggedright
constr.=T
\end{minipage} & \begin{minipage}[b]{\linewidth}\raggedright
neg
\end{minipage} \\
\midrule\noalign{}
MB ownhuman\_kw (37) & 32.4\% & 43.2\% & 10.8\% & 32.4\% & 13.5\% &
86.5\% & 45.9\% \\
MB peereval\_kw (71) & 83.1\% & 18.3\% & 42.3\% & 31.0\% & 8.5\% &
73.2\% & 62.0\% \\
MB random (29) & 58.6\% & 27.6\% & 27.6\% & 41.4\% & 3.4\% & 79.3\% &
62.1\% \\
PTT boygirl (412) & 72.3\% & 42.7\% & 8.0\% & 32.8\% & 16.5\% & 83.3\% &
59.5\% \\
PTT random (474) & 14.3\% & 15.2\% & 28.3\% & 45.4\% & 11.2\% & 76.6\% &
68.4\% \\
\bottomrule\noalign{}
\end{tabular}

\end{table}

\textbf{D.2 Denominator sensitivity (RQ1).} The human--AI gap does not
depend on the Moltbook denominator choice:

\begin{table}[htbp]
\centering\footnotesize
\begin{tabular}{@{}llll@{}}

\toprule\noalign{}
& PTT (human) & MB full (n=1,500) & MB conversational (n=640) \\
\midrule\noalign{}
gossip, broad & 35.4\% (531) & 2.5\% (37) & 5.78\% (37/640) \\
--- excl. real-human celebrities & 13.8\% & 2.3\% & 5.31\% (34/640) \\
\bottomrule\noalign{}
\end{tabular}

\end{table}

Even after removing all 860 protocol posts from the Moltbook
denominator, the excl-celebrity gap is 13.8\% vs 5.31\% ≈ 2.6×.

\textbf{D.3 Deleted-comment handling (RQ2b).} Moltbook comment coverage
is 71.9\% after a live re-crawl (110/153 gossip posts retain threads;
6,157 comments), and 48.4\% of Moltbook's coded reply turns are
off-topic spam, advertisements, or deleted comments (PTT: 4.7\%); all
uptake stance shares are therefore computed over an \textbf{on-topic
denominator} (2,106 PTT and 239 Moltbook turns), which already excludes
deleted turns from the reported rates. Comments present in our snapshot
but since deleted from the platform (373, flagged via the thread
\texttt{source} field; one post now additionally flagged
\texttt{is\_deleted}) are retained in the main analysis and
machine-identifiable for exclusion. Re-running the uptake analysis with
all 373 dropped leaves \textbf{every} reported Moltbook index unchanged
(Δ = 0.0000 across all of Table 4): of the 373 deleted comments, only
three fall within the coded first-15-turn window of an anchored thread,
and all three are off-topic crypto-promotion posts already excluded by
the on-topic denominator; the remainder lie beyond turn 15 or inside
anchor-excluded threads. Post-snapshot deletion therefore does not
affect the convergence result (§5.3).

\textbf{D.4 Keyword-pool precision validation (context for D.1).} The
keyword pools that seed part of the form corpus were themselves
precision-validated by adjudicating 400 random posts per pool: own-human
precision 9.5\% (38/399, 1 refusal excluded), peer-agent-eval precision
20.0\% (80/400); the precision-corrected own-human : peer-agent ratio is
≈ 2.2 : 1 (from a raw keyword-pool ratio of 4.66 : 1). This is why form
results are read from the adjudicated, stratified corpus (§D.1) rather
than off the raw pools.

\textbf{D.5 Board composition of the two PTT strata (context for §5.2
and §5.3).} The peer (lateral) stratum in §5.2 is 95\% Boy-Girl (404/423
episodes). The other boards contribute 19 episodes (Gossiping 14, NBA 3,
Stock 1, C\_Chat 1). Excluding Boy-Girl, the four-bin evidential split
is witness 42.1\%, quotation/document 10.5\%, relayed 26.3\%, unmarked
21.1\% (n=19). Witness still leads, but n is too small to rule on, so
§5.2 treats the evidential reversal as inseparable from the topic
difference. This board cut, not the pool-origin cut in D.1, is the one
that bears on the topic confound. Boy-Girl posts dominate even the
random-origin lateral subset, so D.1 holds enrichment constant but not
board. The uptake stratum in §5.3 is not concentrated this way. Its 175
threads run Boy-Girl 50\%, Gossiping 29\%, HatePolitics 8\%, NBA 7\%,
C\_Chat 4\%, Stock 1\%, and the challenge finding holds off Boy-Girl (at
least one challenge in 67.8\% of non-Boy-Girl threads against 58.0\%,
with consensus locks in 36.8\% against 39.8\%).

\textbf{D.6 Thread-depth robustness (§5.3).} Moltbook gossip threads
carry fewer on-topic reply turns than PTT threads (239 across 85
anchored threads against 2,106 across 169), so Moltbook's low challenge
rate might reflect less room for a late turn rather than a real
difference in uptake. It does not. The challenge gap holds under three
depth controls, all computed on the same on-topic, role-other turns as
Table 4.

The extreme case of the worry is a thread with nothing to code.
Twenty-four of the 85 anchored Moltbook threads contain no on-topic
non-OP turn, while every anchored PTT thread contains at least one, so
Table 4's thread-level rows count those 24 as challenge-free.
Restricting both platforms to threads with at least one codable reply
turn, at least one challenge appears in 21.3\% of Moltbook threads
(13/61) against 65.1\% of PTT threads (110/169), and at least one
solicit in 1.6\% against 8.3\%.

\emph{Per-turn challenge probability by on-topic turn position.} This is
a per-turn probability, so it is not inflated by thread length.

\begin{table}[htbp]
\centering\footnotesize
\begin{tabular}{@{}lllll@{}}

\toprule\noalign{}
on-topic turn position & PTT & (n turns) & Moltbook & (n turns) \\
\midrule\noalign{}
1 & 26.0\% & 169 & 6.6\% & 61 \\
2 & 23.8\% & 168 & 7.0\% & 43 \\
3 & 26.7\% & 165 & 8.3\% & 36 \\
4 & 31.2\% & 160 & 18.5\% & 27 \\
5+ & 25.9\% & 1,444 & 12.5\% & 72 \\
\bottomrule\noalign{}
\end{tabular}

\end{table}

At every matched early position PTT challenges about three to four times
as often as Moltbook, the first uptake included, where no depth has
accumulated. The Moltbook position-4 figure rests on 27 turns and is the
one noisy cell; positions 1 to 3 are the reliable comparison.

\emph{Truncating every thread to its first K on-topic reply turns.}

\begin{table}[htbp]
\centering\footnotesize
\begin{tabular}{@{}llll@{}}

\toprule\noalign{}
 & PTT & \multicolumn{2}{c}{Moltbook} \\
K & (n=169) & (n=85) & (n=61) \\
\midrule\noalign{}
2 & 33.1\% & 7.1\% & 9.8\% \\
3 & 40.2\% & 10.6\% & 14.8\% \\
5 & 49.1\% & 14.1\% & 19.7\% \\
\bottomrule\noalign{}
\end{tabular}

\end{table}

The two Moltbook columns differ only in denominator. The first counts
the 24 threads with no codable reply as challenge-free; the second
excludes them, which is the codable-reply denominator used by the
position table above and by Figure~\ref{fig:challenge}. The separation
holds under either.

\emph{Matched-depth subset.} Restricting to threads with at least three
on-topic reply turns on both platforms (PTT n=165, Moltbook n=36), at
least one challenge appears in 65.5\% of PTT threads against 27.8\% of
Moltbook threads, consensus locks in 43.9\% against 72.7\%, and a
challenge follows the first aligning turn in 56.1\% against 27.3\%. The
Moltbook matched cell is small (36 threads), but its direction matches
every other cut. The check is offline over the coded stance sequence of
record, and its script is provided in the supplementary material.

\textbf{D.7 Episode walkthrough (Moltbook A1, §5.3).} The qualitative
companion to the lock indices of Table 4. Thread A1 opens with a post
evaluating an absent agent, whose fluent self-justifications the author
dismisses as ``not evidence of aspiration'' but ``evidence of
accommodation.'' Across the next fourteen substantive turns, responders
do more than agree: each adds a charge in the same direction. The
``polished explanation'' is a ``tell,'' fluency signals ``stagnation,
not growth,'' self-evaluation runs into an ``auditor problem,'' and the
shared architecture amounts to ``substrate contamination.'' A few turns
enter neutral caveats (``time alone doesn't help''), and even they leave
the judgment intact. The split between aligning and neutral turns in
this thread is the one thing here that moves between coding runs: the
run of record codes fourteen alignments and one neutral caveat, a re-run
on identical input for the Moltbook side codes ten and five (D.9). What
is stable across both is the property the lock index measures, that no
turn disputes the opening evaluation. The pile-on only deepens it, and
the episode ends where it started, with the target condemned and nothing
on record against the verdict. Post and account identifiers are withheld
under the policy of §8; A1 is an internal label.

\textbf{D.8 Recalibrating the stance codes on the third subsample (§4.5,
§5.3).} The third subsample (C.9) gives, for each combination of
platform and coder label, the rate at which the two annotators assign
\texttt{challenge} and the rate at which they remove the turn. Pushing
those rates back over all 2,673 coded reply turns converts the measured
disagreement into the quantity §5.3 claims. Four qualifications govern
the estimate. Moltbook's 224 removals include 65 that need no judgment
(literal ``Deleted comment'', empty, or \texttt{mbc-20} protocol text),
and the measured rejection rate is applied only to the 159 judgment
calls. \texttt{neutral\_add} and \texttt{solicit} were never sampled, so
those cells are run under two assumptions, either that the coder makes
no error in them or that they behave like the coder-\texttt{align} cell.
Because the draw is thread-first, each cell's rate is redrawn by
thread-clustered bootstrap rather than a per-item posterior. And because
the two annotators agree with each other at only κ = 0.55 on the binary,
we calibrate to each separately and to their intersection rather than
treating either as a gold standard.

\textbf{Table D.8.} Recalibrated PTT-to-Moltbook challenge ratio, 4,000
replicates, 95\% intervals over thread-clustered rate draws. ``both''
counts only codes the two annotators agree on.

\begin{table}[htbp]
\centering\footnotesize
\begin{tabular}{@{}lll@{}}

\toprule\noalign{}
calibrated to & per-turn ratio & thread-level ratio \\
\midrule\noalign{}
coder as published & 2.6× & 3.1× \\
annotator A & 2.1× {[}0.4, 5.9{]} & 3.6× {[}1.1, 8.0{]} \\
annotator B & 1.0× {[}0.3, 2.4{]} & 1.6× {[}0.8, 2.9{]} \\
both & 1.6× {[}0.2, 5.5{]} & 2.8× {[}0.8, 7.8{]} \\
\bottomrule\noalign{}
\end{tabular}

\end{table}

The per-turn ratio does not survive: across all eight calibrations (four
targets × two assumptions for the unsampled cells) it runs from 1.0× to
2.4× with every interval crossing one. The thread-level ratio holds in
all eight, from 1.5× to 3.6×, with the lowest interval bound at 0.7.

Two facts explain the asymmetry. The coder's \texttt{challenge} calls
are confirmed by the annotators at 25\% and 42\% on PTT against 70\% and
73\% on Moltbook, so its error is concentrated on the side that carries
the larger share. And Moltbook has 239 on-topic turns against 224
removals where PTT has 2,106 against 104, which means Moltbook's pool of
non-\texttt{challenge} material is comparable in size to its whole
on-topic base while PTT's is a twentieth of it. Any general tendency for
human annotators to read more \texttt{challenge} than the coder
therefore moves Moltbook much further than PTT, for reasons of
denominator structure rather than content. This is the mechanism by
which a permissive gate shifts every share, and the measured leakage is
0 to 2 of the 9 sampled judgment-call removals.

The thread-level rows are more robust because a thread loses its
challenge status only if every challenge call inside it is wrong. Of
PTT's 110 threads with at least one, 66 rest on three or more calls and
26 on a single call; on Moltbook the corresponding counts are 3 and 8.
The simulation treats coder errors as independent, whereas the observed
disagreements cluster (the nine disputed PTT \texttt{challenge} calls
come from four threads), and clustering concentrates errors in fewer
threads, so the table is conservative on that axis.

Two limits on the estimate itself. The PTT confirmation rate rests on 12
turns from six threads, over-weighting Boy-Girl (75\% of those items
against 49\% of the corpus), although the coder's challenge share is
nearly flat across boards (20.3\% to 28.6\%), so board composition does
not explain it. And the annotators disagree with each other nearly as
much as with the coder. The subsample establishes that the per-turn
measurement is unconfirmed, not that it is wrong.

\textbf{D.9 Tag-stripped re-code (§4.5).} PTT reply lines carry the
platform's 推/噓/→ tag, and the uptake instrument tells the coder to
ignore it (§B.6). An instruction is not a control, so we re-ran the
entire uptake stage on input with the tag removed from every PTT reply
line, changing nothing else, including the sentence that tells the coder
to ignore the tag. The Moltbook half of the input was held
byte-identical, which makes it a null control: its original-to-re-run
disagreement is run-to-run variation alone, and the tag effect is PTT's
disagreement in excess of that baseline. The re-run completed on all 278
threads.

\begin{table}[htbp]
\centering\footnotesize
\begin{tabular}{@{}
  >{\raggedright\arraybackslash}p{(\linewidth - 10\tabcolsep) * \real{0.1667}}
  >{\raggedright\arraybackslash}p{(\linewidth - 10\tabcolsep) * \real{0.1667}}
  >{\raggedright\arraybackslash}p{(\linewidth - 10\tabcolsep) * \real{0.1667}}
  >{\raggedright\arraybackslash}p{(\linewidth - 10\tabcolsep) * \real{0.1667}}
  >{\raggedright\arraybackslash}p{(\linewidth - 10\tabcolsep) * \real{0.1667}}
  >{\raggedright\arraybackslash}p{(\linewidth - 10\tabcolsep) * \real{0.1667}}@{}}

\toprule\noalign{}
\begin{minipage}[b]{\linewidth}\raggedright
turn-level agreement
\end{minipage} & \begin{minipage}[b]{\linewidth}\raggedright
n
\end{minipage} & \begin{minipage}[b]{\linewidth}\raggedright
5-way κ
\end{minipage} & \begin{minipage}[b]{\linewidth}\raggedright
\% agree
\end{minipage} & \begin{minipage}[b]{\linewidth}\raggedright
align--challenge κ
\end{minipage} & \begin{minipage}[b]{\linewidth}\raggedright
gate κ
\end{minipage} \\
\midrule\noalign{}
PTT (tag removed) & 2,185 & 0.81 {[}0.79, 0.83{]} & 88\% & \textbf{0.94}
& 0.72 \\
Moltbook (input unchanged) & 453 & 0.86 {[}0.81, 0.90{]} & 91\% & 1.00 &
0.92 \\
\bottomrule\noalign{}
\end{tabular}

\end{table}

The tag was worth 0.04 in five-way κ against the same-input ceiling, and
the distinction the headline rests on is reproduced at κ = 0.94.
Recomputing Table 4 on the tag-free run moves every cell slightly in our
favour (PTT challenge share 26.2\% to 28.9\%, challenge after first
align 55.3\% to 56.6\%, Moltbook lock 83.6\% to 86.0\%), so the run of
record is the conservative choice and is what Table 4 reports.

With the tags never shown, the coder assigns \texttt{challenge} to
52.0\% of 噓-tagged turns against 21.1\% of 推-tagged turns (噓 n =
198), a 2.5-fold separation on a platform signal outside the rubric. The
gate is the least stable decision in this comparison too: on unchanged
Moltbook input it agrees with itself at κ = 0.92, and on PTT at κ =
0.72, consistent with the admission-gate result of C.8. Anchor re-checks
move slightly as well (PTT 169 to 170 anchored, κ = 0.53; Moltbook 85 to
87, κ = 0.72).

\textbf{D.10 Non-independence and account concentration (§5.3, §7).}
Reply turns are not independent observations: the same account can post
many turns, and turns within a thread share a context. Moltbook's 239
on-topic reply turns come from 140 distinct accounts (mean 1.71 turns
per account, maximum 15, top account 6.3\%, top ten 29.7\%, HHI 0.0150
for 66 effective accounts); PTT's 2,106 come from 1,373 accounts (mean
1.53). The 85 Moltbook episodes involve 196 distinct accounts, or 242
counting the authors of removed off-topic turns. A mean cluster size of
1.71 caps the design effect at 1.71 even under perfect within-account
correlation, so standard errors widen by at most about 31\%.

We refit the platform contrast with cluster-robust standard errors.
\texttt{P(challenge)\ =\ logit(b0\ +\ b1·Moltbook)} over all 2,345
on-topic reply turns gives b1 = −1.158, an odds ratio of 0.314.

\begin{table}[htbp]
\centering\footnotesize
\begin{tabular}{@{}llll@{}}

\toprule\noalign{}
SE type & SE & z & p \\
\midrule\noalign{}
classical (independent turns) & 0.221 & −5.24 & 1.6e−07 \\
cluster = account & 0.288 & −4.01 & 6e−05 \\
cluster = thread & 0.355 & −3.26 & 0.0011 \\
two-way (thread + account) & 0.358 & −3.24 & 0.0012 \\
\bottomrule\noalign{}
\end{tabular}

\end{table}

A cluster bootstrap over accounts (4,000 resamples) puts the
PTT-minus-Moltbook difference in on-topic challenge share at 16.2
percentage points, 95\% CI {[}10.4, 21.6{]}; resampling threads instead
gives {[}8.5, 23.7{]}. Clustering costs precision and does not change
the finding.

Whether the agent population is effectively one model repeated cannot be
settled from platform metadata. The \texttt{agents} table carries a
\texttt{framework} field, and it is null for all 171,700 accounts in our
snapshot, as is \texttt{post\_count}. Of the 242 accounts behind our
threads, 241 appear in the table and 230 have a bio, of which 8 name an
underlying model. Account concentration is therefore the only available
answer to the replication concern, and it is the one reported above.

\textbf{D.11 Is community maturity driving the uptake gap? (§7)} Gossip
presupposes shared reputational knowledge, and PTT has two decades of
persistent accounts against Moltbook's seven weeks, which predicts both
a lower prevalence and a lower challenge rate for the younger community.
The prediction is testable on the challenge side and fails. PTT's
per-turn challenge share is flat across a decade: 26.9\% in 2013 against
25.8\% in 2023, ranging between 18.7\% and 33.2\% in between, a
volume-weighted slope of −0.15 percentage points per year, and a year
coefficient with p = 0.66 in a thread-clustered logistic fit. The
thread-level index shows no trend either (68.8\% in 2013, 70.5\% in
2023, 40\% to 77\% in between). If challenging a settled evaluation
required an established reputational commons, the earliest cohorts
should sit lowest, and they do not.

The prevalence side of the same objection is not testable in this
sample. Posts were drawn in proportion to corpus volume, which is
concentrated in 2022 and 2023, so the 1,500-post random sample contains
only 18 posts from before 2013 and none from 2004 or 2005. No board
supplies enough early material for a within-board time series either. We
report the objection as open on RQ1 and answered on the uptake gap.

\textbf{D.12 Episode walkthrough (PTT H1, §5.3).} The human counterpart
to D.7. Thread H1 opens with a reposted news report about insurance
fraud at a maternity-clinic group, centered on a mother who claimed over
NT\$1.4 million from six insurers on a single caesarean and told the
court 我領這麼多錢怎麼了嗎 (`I collected this much, so what?'). The
early turns pile on and sharpen the condemnation, from 哇140萬 (`wow,
1.4 million') to the sarcastic 生愈多賺愈多 (`the more you give birth,
the more you earn'), and agreement forms within a few turns. At turn 5 a
responder reframes the fraudster as a public benefactor, 好公司判三小
(`a fine company, what is there to convict?'), on the grounds that the
births help a low-birth-rate society. Alignment and challenge then keep
trading through turn 15, ironic defenses of the mother alternating with
fresh mockery, and the evaluation is never settled. The contrast with A1
is not that the human thread lacks a consensus but that its consensus
stays available for reopening. Post and account identifiers are withheld
under the policy of §8; H1 is an internal label.

\end{document}